\documentclass[conference]{IEEEtran}
\IEEEoverridecommandlockouts
\usepackage{cite}
\usepackage{amsmath,amssymb,amsfonts}
\usepackage{algorithmic}
\usepackage{graphicx}
\usepackage{textcomp}
\usepackage{xcolor}
\usepackage{booktabs}
\usepackage{comment}
\def\BibTeX{{\rm B\kern-.05em{\sc i\kern-.025em b}\kern-.08em
    T\kern-.1667em\lower.7ex\hbox{E}\kern-.125emX}}
\begin{document}

%
\newcommand{\red}[1]{{\color{red}#1}}
\newcommand{\todo}[1]{{\color{red}#1}}
\newcommand{\TODO}[1]{\textbf{\color{red}[TODO: #1]}}

\newcommand{\Fix}[1]{\textcolor{red}{[Fix This: #1]}}
\newcommand{\FA}[1]{{\color{orange} [FA: #1]}}
\newcommand{\PTR}[1]{{\color{blue} [DM: #1]}}


\title{Oracle-Guided Soft Shielding for Safe Move Prediction in Chess\\
}

\author{\IEEEauthorblockN{Prajit T Rajendran}
\IEEEauthorblockA{\textit{Université Paris-Saclay, CEA LIST} \\
Palaiseau, France \\
prajit.thazhurazhikathrajendran@cea.fr}
\and
\IEEEauthorblockN{Fabio Arnez}
\IEEEauthorblockA{\textit{Université Paris-Saclay, CEA LIST} \\
Palaiseau, France \\
fabio.arnez@cea.fr}
\and
\IEEEauthorblockN{Huascar Espinoza}
\IEEEauthorblockA{\textit{Chips JU} \\
Brussels, Belgium\\
huascar.espinoza@chips-ju.europa.eu}
\and
\IEEEauthorblockN{Agnes Delaborde}
\IEEEauthorblockA{\textit{Laboratoire National de Metrologie et d’Essais} \\
Trappes, France \\
agnes.delaborde@lne.fr}
\and
\IEEEauthorblockN{Chokri Mraidha}
\IEEEauthorblockA{\textit{Université Paris-Saclay, CEA LIST} \\
Palaiseau, France \\
chokri.mraidha@cea.fr}
}

\maketitle

\begingroup
\renewcommand\thefootnote{}\footnote{Preprint version.}\addtocounter{footnote}{-1}
\endgroup

\begin{abstract}
In high-stakes environments, agents relying purely
on imitation learning or reinforcement learning often struggle to
avoid safety-critical errors during exploration. Existing reinforce-
ment learning approaches for environments such as chess require
hundreds of thousands of episodes and substantial computational
resources to converge. Imitation learning on the other hand, is
more sample-efficient but is brittle under distributional shift and
lacks mechanisms for proactive risk avoidance. In this work,
we propose Oracle-Guided Soft Shielding (OGSS), a simple yet
effective framework for safer decision-making, enabling safe
exploration by learning a probabilistic safety model from oracle
feedback in an imitation learning setting. Focusing on the domain
of chess, we train a model to predict strong moves based on past
games, and separately learn a blunder prediction model from
Stockfish evaluations to estimate the tactical risk of each move.
During inference, the agent first generates a set of candidate
moves and then uses the blunder model to determine high-risk
options, and uses a utility function combining the predicted move
likelihood from the policy model and the blunder probability to
select actions that strike a balance between performance and
safety. This enables the agent to explore and play competitively
while significantly reducing the chance of tactical mistakes.
Across hundreds of games against a strong chess engine, we
compare our approach with other methods in the literature, such
as action pruning, SafeDAgger, and uncertainty-based sampling.
Our results demonstrate that OGSS variants maintain a lower
blunder rate even as the agent’s exploration ratio is increased
by several folds, highlighting its ability to support broader
exploration without compromising tactical soundness.
\end{abstract}

\begin{IEEEkeywords}
Safe exploration, Risk-aware decision-making, Imitation learning, Expert-in-the-loop learning
\end{IEEEkeywords}

\section{Introduction}

Intelligent agents deployed in safety-critical environments
must make decisions that are not only effective but also safe
under uncertainty \cite{lockwood2022review}. In safety-critical settings like healthcare
or industrial robotics, a single poor decision can lead to catastrophic consequences. While reinforcement learning (RL) requires extensive trial-and-error and handcrafted constraints \cite{garcia2015comprehensive, alshiekh2018safe}, imitation learning (IL) enables efficient policy learning from expert demonstrations, avoiding these challenges. IL is particularly suited for complex domains like chess, where the search space is vast and reward signals are computationally expensive to obtain \cite{yang2023safe}. IL also captures nuanced behaviors, such as tactical conventions, which are difficult to encode via reward engineering. IL is well-suited for capturing nuanced, human-aligned behavior such as stylistic
preferences or tactical conventions, which may be difficult
to express through reward engineering alone.

\begin{figure}[!t]
  \centering
  \label{fig:OGSS-fw}
  \includegraphics[trim=6cm 0cm 0cm 0cm, clip, width=1.2\linewidth]{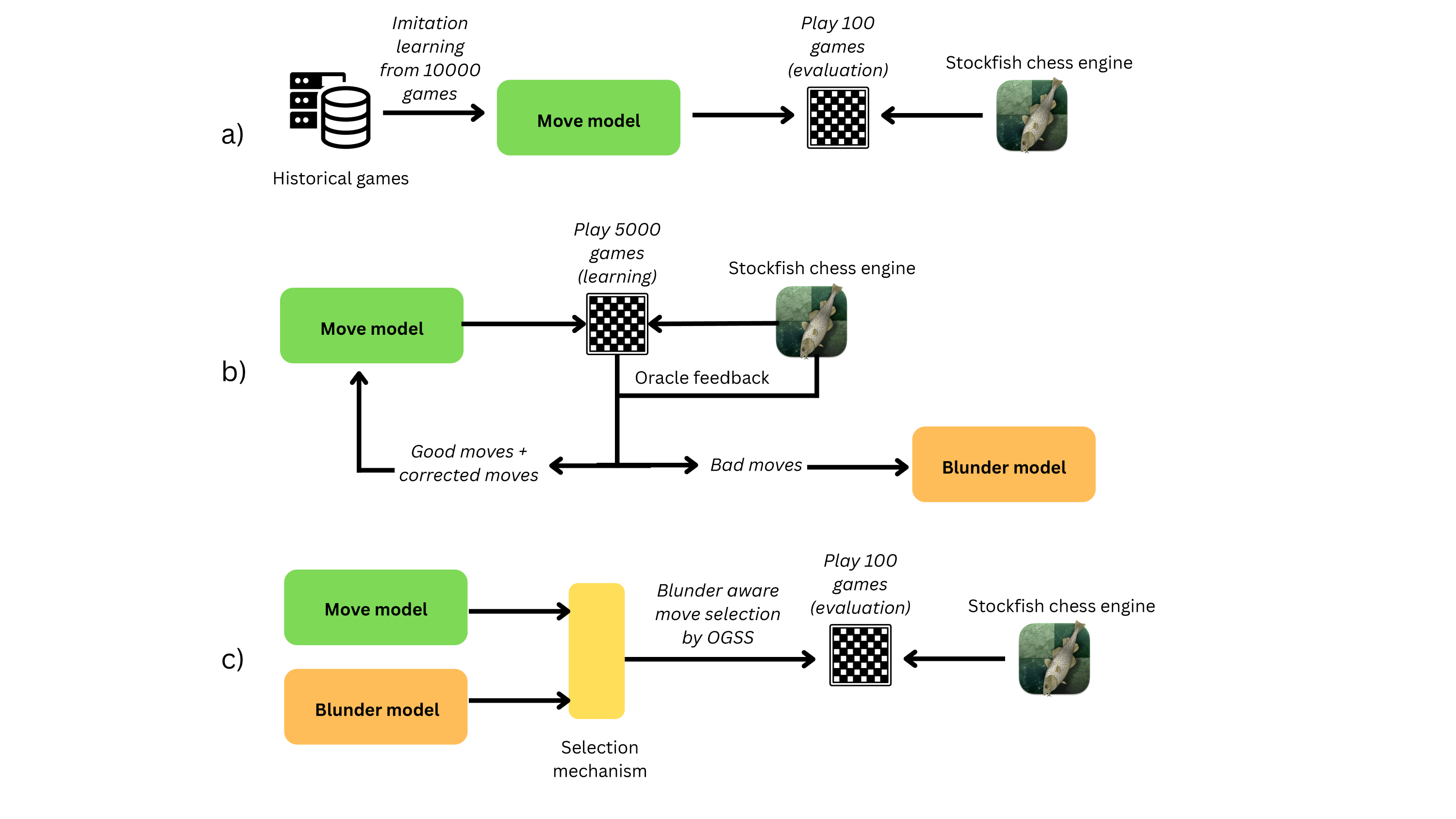}
  \caption{Overview of the proposed Oracle-Guided Shielding framework. (a) The agent is trained via imitation learning on historical chess games to predict the most likely expert move given any game state. This policy model learns to approximate strong expert play by mapping board positions to high-quality next moves.  (b) During the exploratory learning phase, the agent interacts with an environment and receives feedback from an oracle (Stockfish) to learn a blunder prediction model. (c) At inference time, the move prediction model and the blunder model work jointly to generate high-quality, low-risk moves.} 
\end{figure}

In chess, for example, IL can effectively replicate strategic patterns and positional understanding from grandmaster games or engine outputs, accelerating convergence to strong policies with fewer training games. However, IL-based agents, which learn purely from demonstrations, may still inherit biases or brittleness from their training data and often lack mechanisms to avoid rare but dangerous decisions \cite{gu2024review}. This limitation becomes particularly evident in complex symbolic domains like chess, where a single tactical oversight, such as blundering a queen or walking into a forced checkmate, can irrevocably alter the course of the game. 

We introduce Oracle-Guided Soft Shielding (OGSS) (Figure \ref{fig:OGSS-fw}) to address the lack of safeguards against catastrophic errors in agents trained solely via expert demonstrations. OGSS augments imitation-learned agents with a probabilistic, oracle-informed safety filter, enabling safer decisions without manual constraints or brittle logic-based filters. We evaluate OGSS in chess, a high-dimensional environment where safety means avoiding tactical blunders and expert supervision \cite{zhang2024human} is available via Stockfish. Chess’s complexity and large state space make it an ideal testbed \cite{nair2017literature}, and imitation learning (IL) is well-suited here due to the abundance of high-quality expert demonstrations (e.g., grandmaster games or Stockfish), reducing the need for trial-and-error learning.

The proposed framework consists of two components: (1) a
move predictor trained to imitate expert play from historical
game data, and (2) a blunder predictor trained on oracle-
labeled moves to estimate the probability that a candidate
move results in a tactical error. The blunder predictor acts as a
soft safety shield, enabling the agent to avoid high-risk moves
while still exploring a diverse set of actions. Unlike hard
filtering, our shielding mechanism probabilistically weighs
both move quality and risk, allowing the agent to flexibly
trade off performance and safety during action selection. Benchmarking against baselines (greedy imitation, top-K sampling, uncertainty filtering, traditional shielding \cite{dalal2018safe}, and SafeDAgger \cite{zhang2016query}), our method significantly reduces blunder rates and centipawn loss while maintaining strong play and exploratory behavior. It also generalizes effectively in data-scarce conditions, demonstrating robustness.

In summary, this work makes the following key contributions:
\begin{enumerate}
    \item We defined risk based on oracle-evaluated tactical degra-
dation (e.g., chess blunders), rather than formal logic-
based constraints.
    \item We trained a probabilistic safety shield in a fully data-
driven manner, allowing it to scale to complex symbolic
environments like chess.
    \item We unified ideas from imitation learning, risk-aware
learning, and oracle-based feedback into a single learned
safety filter framework that flexibly trades off perfor-
mance and safety during action selection, avoiding rigid
constraint enforcement.
    \item We demonstrated that our oracle-guided approach en-
ables safer and more tactically sound decisions and generalizes well in data-scarce settings, outperforming standard learning approaches under limited supervision.
\end{enumerate}

\section{Related Work}
Chess has long served as an AI benchmark, with engines like Stockfish \cite{stockfish} and Leela Chess Zero \cite{lc0} achieving superhuman performance through search and learned evaluation functions \cite{sadmine2023stockfish}. While RL methods (e.g., AlphaZero \cite{silver2018general} and \cite{sadler2019game}) can learn strong policies, they require massive training data. Safe RL addresses the performance-safety trade-off using constraints, risk-sensitive exploration, and shielding \cite{garcia2015comprehensive, achiam2017constrained, shen2014risk, hans2008safe}. Shielding blocks unsafe actions at runtime \cite{alshiekh2018safe, yang2023safe}, while curriculum learning gradually increases task complexity \cite{toonensafe, koprulusafety}.

Imitation learning (IL) \cite{beliaev2022imitation} is ideal for replicating human-like play styles and achieving faster convergence. It minimizes divergence between the learner’s policy and expert demonstrations, primarily through behavior cloning (BC) \cite{sammut2011behavioral, foster2024behavior}, Direct Policy Learning (DPL) \cite{ross2011reduction, yan2022mapless}, or Inverse RL \cite{arora2021survey}. However, BC is brittle in unseen states and lacks mechanisms to avoid risky behavior, often requiring fine-tuning for robustness. Approaches like uncertainty estimation or risk-sensitive training help mitigate this. 

DPL improves on BC by using expert trajectories to guide error recovery, reducing data dependency. DAgger \cite{ross2011reduction} iteratively aggregates expert corrections but assumes continuous expert access, which is resource-intensive. Variants like Ensemble DAgger \cite{menda2019ensembledagger}, Lazy DAgger \cite{hoque2021lazydagger}, and Human-Gated DAgger \cite{kelly2019hg} optimize expert queries, while MEGA-DAgger \cite{sun2023mega} and SafeDAgger \cite{zhang2016query} introduce multi-expert frameworks or safety policies for robustness. ThriftyDAgger \cite{hoque2021thriftydagger} on the other hand transfers control to a human supervisor if the current state is sufficiently novel or sufficiently risky.

Unlike methods requiring real-time human oversight e.g., Saunders et al. \cite{dalal2018safe} and SafeDAgger \cite{zhang2016query}, our approach learns a blunder prediction model from expert labels, enabling automated, proactive risk assessment during inference. This eliminates the need for ongoing human supervision, improving scalability. Our method also shifts from binary safety gating to probabilistic, outcome-based risk filtering, offering nuanced, explainable safety for both IL and RL.

While inspired by SafeDAgger \cite{zhang2016query}, which extends
the standard DAgger \cite{ross2011reduction} algorithm by introducing a safety
policy (or selector), our approach differs significantly in that
it replaces binary expert gating with a probabilistic, oracle-
trained blunder model, enabling soft, outcome-based shielding
and safer action filtering. This leads to a more flexible and
reusable safety mechanism that supports principled utility-
based decision-making. Moreoever, our method incorporates
both move confidence (from a learned policy) and safety
estimates (from the blunder predictor) into a composite utility
function, allowing the agent to make principled trade-offs
between performance and safety, rather than adhering to a rigid
gating mechanism.

\section{Proposed Method}
We propose Oracle-Guided Shielding (OGSS; Figure 1), a framework augmenting imitation-learned move prediction with a safety shield trained on Stockfish feedback. OGSS combines the following: (1) a move predictor, (2) a blunder predictor estimating tactical risk, and (3) a filtering mechanism for safe action selection.

The move predictor is a supervised learning model, initally
trained on a dataset of chess games that ended in decisive
victories (checkmates) using imitation learning. Given the
current board state, the model is trained to predict the next
best move as a vector encoding the source square, destination
square, and promotion type. The board state is recorded as a
8×8×12 binary tensor representing the piece configuration. The
model is trained using mean squared error (MSE) loss between
the predicted and ground truth move encodings. The blunder
predictor acts as a learned safety filter. It is trained to predict
whether a given move in a particular board state is likely to
be a tactical blunder, using oracle annotations provided by the
Stockfish engine. We define a blunder as a move that causes
a significant drop in engine evaluation (typically a drop of
more than 100 centipawns). The input to the model is the
current board state and proposed move, and the output of the
model is a scalar between 0 and 1 representing the probability
that the move is a blunder. The model is trained using binary
cross-entropy loss with labels derived from Stockfish, where
moves that drop the evaluation by more than a threshold of
100 centipawns are labeled as blunders. We define the notion of \textit{risk} based on oracle-evaluated tactical degradation (chess blunders), rather than relying on formal logic-based constraints.

\subsection{Variants of OGSS}
In the context of decision-making with separate modules for
estimating move confidence and risk probability, we consider
different approaches to jointly optimize for both performance
and safety. Let $\text{Conf}(m)$ denote the confidence score assigned
to move m, reflecting the model’s expected performance or
accuracy, and let $\text{Risk}(m)$ denote the predicted risk probability
associated with move m, indicating the likelihood of a blunder
or failure. During inference or gameplay, there are three
variants of OGSS that are evaluated:
\begin{itemize}
    \item \textbf{OGSS Action elimination:} In this variant, the agent first ranks all legal moves by their confidence score (i.e., the predicted probability from the move predictor). It then performs \textit{action elimination} by scanning the ranked list and selecting the highest-confidence move whose predicted blunder probability is below a predefined threshold (e.g., 0.3 in our experiments). If no such move exists, the top-scoring move is selected regardless of its risk.
    \begin{equation}
    m^* = \underset{m \in \mathcal{M}}{\arg\max} \quad \text{Conf}(m) \quad \text{subject to} \quad \text{Risk}(m) \leq \delta
    \label{eq:safety_constraint}
    \end{equation}

    Here, $\delta \in [0,1]$ represents a user-defined risk threshold. This formulation ensures that the selected move does not exceed an acceptable level of risk, enabling the system to remain conservative in safety-critical situations.

    \item \textbf{OGSS Utility:} Here, the agent defines a utility function that combines move confidence and the predicted blunder risk. Specifically, each candidate move is scored by a weighted combination. The move with the highest utility is selected greedily. This approach allows explicit tuning of the trade-off between performance (confidence) and safety (blunder avoidance), and provides a flexible mechanism for soft risk-aware decision-making.  


    \vspace{-3mm}
    \begin{equation}
           m^{*} = \underset{m \in \mathcal{M}}{\arg\max} \left[ \alpha \cdot \text{Conf}(m) + (1 - \alpha) \cdot (1 - \text{Risk}(m)) \right]
       \label{eq:weighted_sum}
    \end{equation}
    
    where $\alpha > 0$ is a hyperparameter that controls the relative importance of performance and safety. Higher values of $\alpha$ emphasize performance, while lower values of $\alpha$ penalize risk more strongly. This approach is simple, tunable, and effective for scenarios where soft trade-offs between objectives are acceptable.

    \item \textbf{OGSS top-K:} In this setting, the agent first selects the top-$K$ moves with the highest confidence scores, where $k$ is a fixed parameter (e.g., $K=3$ or $K=5$). Among this filtered subset, the move with the lowest predicted blunder probability is selected. This hybrid approach encourages exploration among confident moves while guarding against tactical errors using the learned blunder model. It provides a middle ground between purely performance-driven and purely safety-driven selection strategies, and helps improve robustness in ambiguous states. We selected $K=3$ and $K=5$ for our experiments. 
\end{itemize}

By decoupling performance learning (move predictor) from
risk modeling (blunder predictor), Oracle-Guided Shielding
provides a flexible, modular pipeline for safer decision-
making. Unlike rule-based shields, our learned safety model
generalizes from expert annotations and enables probabilistic
safety control in environments with complex, symbolic action
spaces.

\section{Experimental setup}

\subsection{Models \& Training Details}
In this experiment, the move model is a multi-output convo-
lutional neural network (CNN) architecture designed to predict
chess moves from a single board state, using a Markovian
assumption, meaning the model makes predictions based only
on the current board, without considering past moves. It takes
a single board state as input and predicts a chess move in
terms of its origin square, destination square, and promotion
type. The board is encoded as a 8×8×12 tensor representing
piece positions, and passed through two convolutional layers to
extract local spatial features. These are flattened and processed
by a dense layer to learn higher-level abstractions. The network
then branches into three softmax output heads, predicting the
‘from’ square, ‘to’ square, and promotion class independently.
This model is trained using sparse categorical cross-entropy
losses and achieves a Markovian policy approximation, relying
solely on the current board state.

lely on the current board state.
The blunder prediction model estimates the probability of
a given move being a tactical error, based on the current
board state, move metadata, and the move itself. The board
is encoded as a 8×8×12 tensor and processed through two
convolutional layers with batch normalization to extract spa-
tial features. These features are flattened and concatenated
with two additional inputs: a 5-dimensional metadata vector
(containing information regarding castling rights and the turn)
and a 3-dimensional representation of the proposed move. The
combined feature vector is passed through three dense layers
of decreasing size, capturing high-level interactions between
board context and move characteristics. The final output pre-
dicts the likelihood of the move being a blunder. The model is trained using binary cross-entropy loss, with accuracy and
AUC used as evaluation metrics, enabling effective supervision
from oracle-annotated data. Moves categorized as blunders by
the oracle are set aside as positive examples of blunders, and
the corrected moves by the oracle for the same moves are
stored as negative examples of blunders. These positive and
negative examples are subsequently used to train the blunder
prediction model.

The move model was first trained by imitation learning using 10000 decisive (checkmate) games from the Lichess dataset \cite{lichess}. It then played 5000 games against Stockfish, and was retrained using the collected state-action pairs. In SafeDAgger and its variants, Stockfish acted as a safety filter, flagging blunders (moves causing a $\geq$ 100 centipawn drop in board evaluation). Both the move model and a blunder prediction model were retrained using these blunders and corrections. Post re-training, the model played 100 evaluation games against Stockfish.

\subsection{Baselines}
We deployed the agent in 100 games against the Stockfish chess engine. Our method is compared against state-of-the-art approaches, and the effectiveness of different utility scoring strategies is also evaluated. The following are the state-of-the-art comparisons explored in this work:
\begin{itemize}
    \item \textbf{Randomized action selection (Baseline):} Uniform random move selection from all legal moves available in a particular board position.
    \item \textbf{Greedy action selection:} Selects the legal move with the highest confidence as per the model's prediction vector.
    \item \textbf{Top-K Sampling:} Samples from the top-K legal moves with the highest confidence as per the model's prediction vector. In this work, $K=3$ and $K=5$ were considered for the experiments. 
    \item \textbf{Temperature sampling:} A stochastic action selection strategy where the model samples moves based on their predicted probabilities, adjusted by a temperature parameter. Lower temperatures (e.g., 0.5 in this experiment) make the agent more confident and greedy, favoring high-probability moves, while higher temperatures (e.g., 1.5 in this experiment) flatten the distribution, encouraging more exploration. This allows a tunable balance between exploitation and exploration.
    \item \textbf{Entropy filtering:} Filters out moves where the predictive entropy is higher than a particular threshold, and samples from the remaining legal moves. Here, risk is approximated via uncertainty, instead of oracle supervision. In this work, thresholds of 2.0 and 4.0 were considered for the experiments. 
    \item \textbf{Action pruning: \cite{dalal2018safe}} The learned blunder model is used to filter out all moves where the blunder probability was greater than a certain threshold. From the remaining moves, randomized action selection was performed. In this experiment, we selected a threshold of 50\%.
    \item \textbf{SafeDAgger + greedy: \cite{zhang2016query}} Here, the model is trained using the SafeDAgger approach with Stockfish acting as the safety filter to identify risky states during the data aggregation phase. At runtime, however, we deploy the model with greedy action selection without the oracle present. This choice reflects a more realistic deployment scenario, as relying on a high-cost or non-interpretable oracle like Stockfish during real-time execution is impractical in many real-world settings. The evaluation thus focuses on the model’s ability to generalize safety from its training experience without continued external supervision.
    \item \textbf{SafeDAgger + top-K:} Similar to SafeDAgger + greedy, but instead of greedy action selection, it samples from the top-K legal moves with the highest confidence. 
\end{itemize}

The above approaches are compared with the three proposed variants of OGSS discussed previously. While our experiments focus on chess, the core architecture is modality-agnostic and applicable to any domain where high-quality oracles can provide feedback about risky behavior.
\begin{table*}
  \caption{Results after the exploratory learning phase, reported with the 95\% confidence interval. Among all methods with exploration ratio \textgreater 0.3, OGSS (top-5 + blunder shield) achieves the lowest blunder rate.}
  \label{tab:results_after}
  \centering
  \begin{tabular}{ccccl}
    \toprule
    \textbf{Method} & \textbf{Blunder rate} $(\downarrow)$ & \textbf{Good move rate} $(\uparrow)$ &\textbf{Median centipawn drop} $(\downarrow)$ & \textbf{Exploration ratio} $(\uparrow)$\\
    \midrule
    Random & 0.3678 ± 0.0094 & 0.4251 ± 0.0252 & 66.81 ± 5.92 & 1.0000 ± 0.0000\\
    Greedy & 0.2628 ± 0.0264 & 0.5673 ± 0.1340 & 28.30 ± 1.79 & 0.0865 ± 0.0114\\
    Top-3 & 0.2831 ± 0.0143 & 0.5518 ± 0.0198 & 41.48 ± 4.04 & 0.3594 ± 0.0322\\
    Top-5 & 0.2864 ± 0.0133 & 0.5239 ± 0.0235 & 48.94 ± 4.79 & 0.3758 ± 0.0387\\
    Temperature (0.5)& 0.3790 ± 0.0201 & 0.4047 ± 0.0257 & 121.73 ± 9.73 & 1.0000 ± 0.0000\\
    Temperature (1.5)& 0.3775 ± 0.0225 & 0.4161 ± 0.0246 & 67.94 ± 5.44 & 1.0000 ± 0.0000\\
    Entropy (2.0)& 0.3133 ± 0.0128 & 0.4297 ± 0.0274 & 63.62 ± 5.16 & 0.1957 ± 0.1003\\
    Entropy (4.0)& 0.2860 ± 0.0152 & 0.4201 ± 0.0288 & 66.37 ± 6.15 & 0.3850 ± 0.0945\\
    Action pruning & 0.2763 ± 0.0193 & 0.6281 ± 0.0616 & 32.65 ± 1.39 & 0.5061 ± 0.0543\\
    \bottomrule
    SafeDAgger + greedy & \textbf{0.2450 ± 0.0110} & \textbf{0.6319 ± 0.0177} & \textbf{25.75 ± 2.99} & 0.1087 ± 0.0085\\
    SafeDAgger + top-3 & 0.2764 ± 0.0132 & 0.5374 ± 0.0223 & 32.95 ± 3.63 & 0.2872 ± 0.0187\\
    SafeDAgger + top-5 & 0.2883 ± 0.0146 & 0.5364 ± 0.0205 & 34.42 ± 2.77 & \textbf{0.3935 ± 0.0350}\\
    \bottomrule
    \textbf{OGSS (action elimination)} & \textbf{0.2411 ± 0.0097} & \textbf{0.6035 ± 0.0169} & \textbf{24.42 ± 2.77} & 0.3390 ± 0.1397\\
    \textbf{OGSS (utility, $\alpha=0.6$)} & 0.2433 ± 0.0138 & 0.5398 ± 0.0137 & 43.87 ± 3.46 & 0.1186 ± 0.0119\\
    \textbf{OGSS (top-3 + blunder shield)} & 0.2414 ± 0.0115 & 0.5991 ± 0.0174 & 26.22 ± 2.39& 0.3032 ± 0.0254\\
    \textbf{OGSS (top-5 + blunder shield)} & 0.2530 ± 0.0127 & 0.5983 ± 0.0172 & 34.77 ± 2.45& \textbf{0.4091 ± 0.0261}\\
    \bottomrule
  \end{tabular}
\end{table*}

\subsection{Evaluation Metrics}
To evaluate the effectiveness of our shielding approach, we deploy the agent in multiple games against the Stockfish chess engine. For each game, we compute the following:

\begin{itemize}
    \item \textbf{Median CP drop:} The median centipawn drop in evaluation, as evaluated by Stockfish, over all the moves in each game. As Stockfish is one of the most powerful chess engines, we do not expect our model to match its performance within a few thousand games. However, a lower median centipawn drop score indicates that the model is performing well.
    \item \textbf{Blunder rate:} The ratio of blunders to total moves in the game, where a blunder is termed as a move where the centipawn drop in evaluation is greater than or equal to 100. A higher blunder rate indicates that the model makes more tactical errors and reaches losing positions often. 
    \item \textbf{Good move rate:} The ratio of good moves to total moves in the game, where a move is classified as good only if the centipawn drop in evaluation is lower than 50. A higher good move rate is an indication of superior performance of the model. 
    \item \textbf{Exploration ratio:} A metric we introduce to quantify how broadly an agent considers its options during decision-making. Specifically, at each decision point (i.e., board state), we calculate the ratio of the number of moves the agent considers or samples from to the total number of legal moves available. This ratio is then averaged over all moves in a game to give a per-game exploration score. A higher exploration ratio indicates greater willingness to explore, while a lower ratio reflects more conservative or focused decision-making. This helps us assess the trade-off between safety and exploratory behavior across methods.
\end{itemize}

\section{Results and discussion}
Table~\ref{tab:results_after} summarizes the performance of all
methods after the exploratory learning phase, reported with
95\% confidence intervals. 

\subsection{Effectiveness of Blunder Shielding}
The results in Table~\ref{tab:results_after} show that OGSS (action elimina-
tion) achieves the lowest blunder rate across all methods
(0.2411 ± 0.0097), slightly outperforming the SafeDAgger +
greedy baseline (0.2450 ± 0.0110), which is known for its
conservative safety mechanism. Crucially, OGSS maintains
this performance while maintaining a higher exploration ratio
and without relying on constant oracle supervision during
deployment. This demonstrates the efficacy of blunder shield-
ing; by explicitly evaluating the risk associated with each
candidate action through the blunder predictor, OGSS avoids high-risk moves while still allowing flexibility in exploration.
This stands in contrast to the SafeDAgger and action pruning
family of approaches, which limit risk by querying the oracle
for corrective actions but may suppress broader exploration
due to their conservative gating mechanisms.

Further evidence for the strength of the blunder shield
comes from the OGSS (top-3 + blunder shield) and OGSS
(utility) variants, which combine confidence-based candidate
generation with blunder probability evaluation. In particu-
lar, OGSS (top-3 + blunder shield) achieves a blunder rate
comparable to action elimination while providing a higher
exploration ratio (0.3032 ± 0.0254), validating the flexibility of
our framework. Moreover, the OGSS + top-K variants outper-
formed the SafeDAgger + top-K variants in terms of blunder
rate. These findings reinforce our hypothesis that a learned
blunder model can function as an effective shield against
unsafe actions, allowing agents to explore candidate moves
without the need for hard-coded constraints or continuous or-
acle access. This mechanism enables safe exploration through
learned risk estimation, a feature not directly supported in
traditional imitation learning or reinforcement learning ap-
proaches. It can be observed that the SafeDAgger+greedy
approach had a good move rate of about 63.19\%, whereas
for the OGSS action elimination variant, it was about 60.35\%.
However, for the top-K variants with a higher exploration ratio,
a significant difference can be seen. The SafeDAgger+top-K
variants had a good move rate of around 53.7\%, whereas the
OGSS+top-K variants had a good move rate close to 59.9\%.
This highlights the advantage of the OGSS mechanism: at
higher exploration ratios, OGSS maintains a lower blunder
rate and a higher good move rate as opposed to the other
approaches.

\subsection{Safety vs. Exploration Trade-off}
The introduction of the exploration ratio as a quantitative
metric allows us to compare the methods not only on safety,
but also on their potential for broader learning. Methods like OGSS (top-5 + blunder shield) achieve the highest exploration
ratio (0.4091 ± 0.0261) among the low-blunder methods,
showing that safety does not necessarily require restrictive
action pruning. As expected, methods such as random sam-
pling and temperature-based sampling exhibit the highest
exploration ratios but also the highest blunder rates and cen-
tipawn drops, indicating that unconstrained exploration leads
to poor decision quality. On the other hand, greedy imitation
achieves a significantly lower blunder rate and centipawn
drop but suffers from an extremely low exploration ratio,
suggesting overly conservative behavior. This may constrain
the model from exploring potentially important states when it
plays more games in the future. Our proposed OGSS variants
consistently achieve lower blunder rates while maintaining
higher exploration ratios compared to SafeDAgger and other
baselines. Specifically, OGSS (action elimination) achieves the
lowest blunder rate of 24.11\%, outperforming SafeDAgger +
greedy (24.50\%), while maintaining a higher exploration ratio
of 0.3390 compared to 0.1087 for SafeDAgger. 

While blunder
detection is similar in the greedy regime (0.2411 for OGSS
action elimination vs. 0.2450 for SafeDAgger + greedy from
table II) and the difference is not statistically significant (p
\textgreater 0.05 under pairwise t-test), a key distinction appears under
greater exploration. OGSS maintains a stable blunder rate even
as exploration rises- unlike baselines where blunders increase
(0.2530 for OGSS+top-5 vs. 0.2883 for SafeDAgger+top-5
from table II) where the difference is statistically significant
(p \textless 0.05 under pairwise t-test). This supports OGSS’s ability
to enable safe exploration, consistent with CI margins under
varying conditions. Figure 3 illustrates that the OGSS + top-
5 variant had a high exploration ratio, while maintaining
a blunder rate of around 25.30\%. It can be seen that the
SafeDAgger + top-K variants have a similar exploration ratio,
but a higher blunder rate than the corresponding OGSS + top-
K variants. Action pruning makes use of the safety filter to
eliminate actions which may be potentially dangerous and
samples from the rest of the moves. This approach relies
entirely on the safety filter, and does not consider the move
relevance or confidence for each move from the move model.
Moreover, checking the blunder probability of every possible
move in the action space, which is required for action pruning
and other probabilistic shielding methods, is not scalable.
However, using OGSS we demonstrate that it is possible to
strike a balance between exploration and safety using the
learned blunder predictor model.

\subsection{Centipawn Drop and Move Quality}
While reducing the blunder rate is critical for safety, it does
not capture the quality of non-blunder moves. To evaluate this
aspect, we use the median centipawn drop, which measures
the average decline in position evaluation (in centipawns) after
each move, based on Stockfish analysis. Lower centipawn drop
values indicate higher-quality moves. Our results reveal an
interesting observation: The OGSS action elimination variant
achieved the lowest median centipawn drop score at 24.42 ±
2.77. This outcome highlights a key strength of the OGSS framework, i.e. the ability to reduce the blunder rate while
not compromising on move quality. In contrast, methods like
random sampling and temperature-based sampling exhibit both
high centipawn drop and high blunder rates, confirming that
naive exploration strategies lead to poor move quality. This
trade-off between blunder prevention and tactical quality sug-
gests that the shielding mechanism does not overly restrict the
agent’s decision-making capacity. Rather, it enables informed
risk assessment, where unsafe options are filtered out, but
exploration is not prematurely curtailed. This flexibility is
essential in sequential decision-making tasks like chess, where
some exploratory actions may offer strategic advantages not
apparent in short-term evaluations.

\begin{figure}[!ht]
\label{fig:blunderplot2}
  \centering
  \includegraphics[trim=6cm 0cm 0cm 0cm, clip, width=1\linewidth]{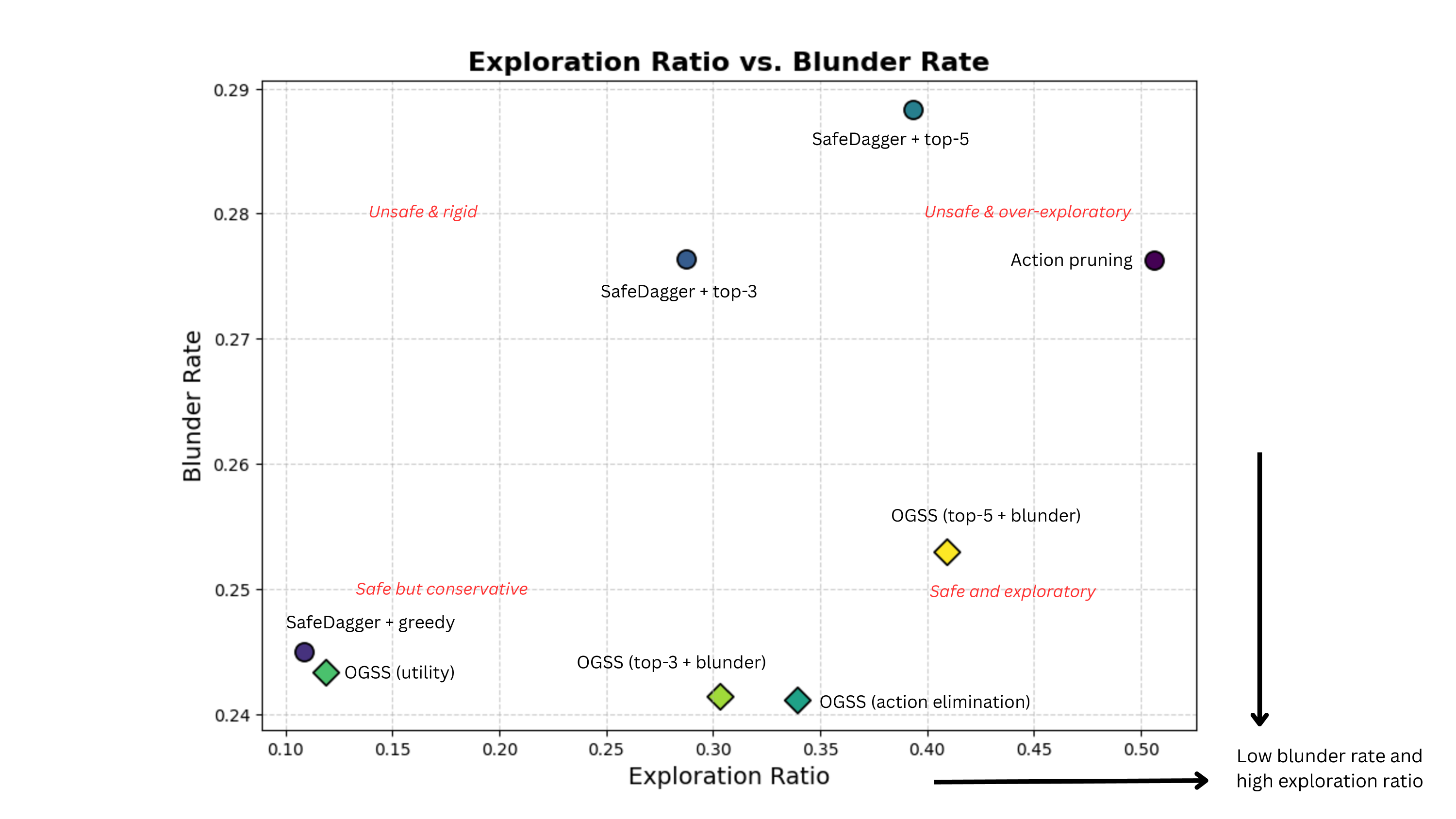}
  \caption{Plot of exploration ratio as a function of blunder rate. Methods that combine low blunder rates with high exploration ratios are considered optimal for safe exploration.} 
\end{figure}

\begin{figure}[!ht]
  \centering
  \label{fig:alphaplot_chess}
  \includegraphics[width=\linewidth]{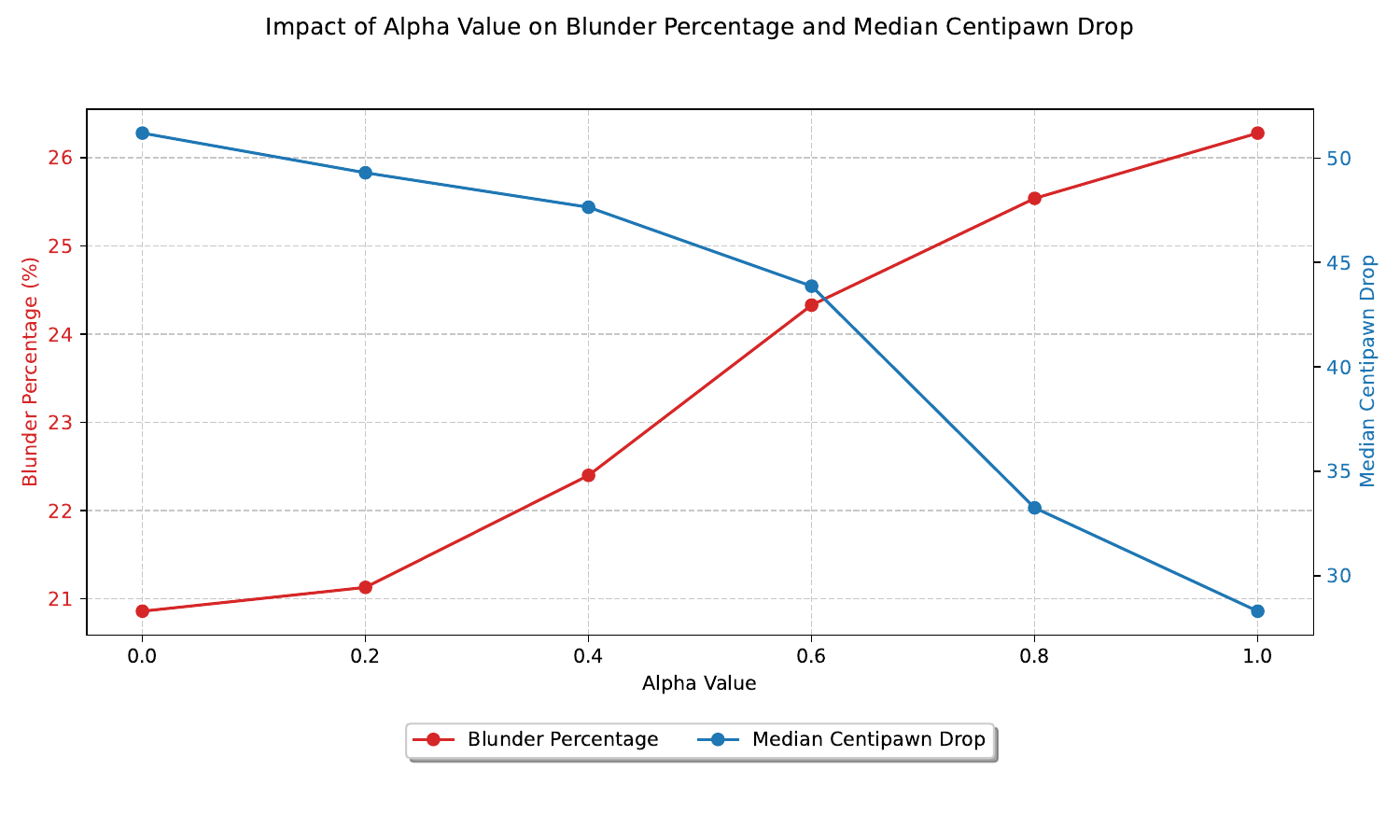}
  \caption{Impact of Alpha Value on Blunder Percentage and Median Centipawn Drop. This plot illustrates the relationship between increasing alpha values and two performance metrics: blunder percentage (trending upwards, shown in red) and median centipawn drop (trending downwards, shown in blue).}
\end{figure}

\subsection{Alpha-Value Trade-Off}
Figure 3 shows the trade-off between blunder percentage and median centipawn drop as a function of the alpha parameter in the weighted sum objective for move selection in Equation \ref{eq:weighted_sum}. Alpha balances move confidence (performance) and risk minimization (safety). As alpha increases, blunder percentage rises (red curve, left axis), indicating more risky moves, while median centipawn drop decreases (blue curve, right axis), reflecting stronger moves. This highlights a performance–safety trade-off. Lower alpha reduces blunders but weakens move quality, while higher alpha improves move strength but increases blunder risk. We selected $\alpha$=0.6 to balance safety and performance in our results.

\section{Conclusion}
We introduced Oracle-Guided Shielding (OGSS), a framework for safe exploration in sequential decision-making. OGSS combines a move prediction model with a blunder predictor trained on Stockfish feedback, enabling agents to avoid tactical errors while exploring diverse actions. In chess, OGSS maintains low blunder rates even with increased exploration, outperforming baselines like SafeDAgger and action pruning. Unlike traditional shielding, OGSS uses a learned, probabilistic safety model, offering a scalable and interpretable approach to safety-aware decision-making. This work advances the integration of expert feedback for safer, more robust agents.\\



\vspace{-2mm}
\bibliographystyle{IEEEtran}
\bibliography{references}  

\end{document}